\documentclass[conference]{IEEEtran}
\IEEEoverridecommandlockouts
\usepackage{cite}
\usepackage{amsmath,amssymb,amsfonts}
\usepackage{algorithmic}
\usepackage{graphicx}
\usepackage{textcomp}
\usepackage{xcolor}
\usepackage{pgfplots}
\usepackage{booktabs}  
\pgfplotsset{compat=1.18} 

\usepackage{soul, color}
\usepackage{colortbl}
\definecolor{Gray}{gray}{0.9}

\def\BibTeX{{\rm B\kern-.05em{\sc i\kern-.025em b}\kern-.08em
    T\kern-.1667em\lower.7ex\hbox{E}\kern-.125emX}}
\begin{document}

\title{Dynamic Gesture Recognition in Ultra-Range Distance for Effective Human-Robot Interaction\\
\thanks{}
}

\author{\IEEEauthorblockN{1\textsuperscript{st} Eran Bamani Beeri}
\IEEEauthorblockA{\textit{School of Mechanical Engineering} \\
\textit{Tel-Aviv University}\\
Tel-Aviv, Israel 6997801 \\
eranbamani@mail.tau.ac.il}
\and
\IEEEauthorblockN{2\textsuperscript{nd} Eden Nissinman}
\IEEEauthorblockA{\textit{School of Mechanical Engineering} \\
\textit{Tel-Aviv University}\\
Tel-Aviv, Israel 6997801 \\
edennissinman@gmail.com}
\and
\IEEEauthorblockN{3\textsuperscript{rd} Avishai Sintov}
\IEEEauthorblockA{\textit{School of Mechanical Engineering} \\
\textit{Tel-Aviv University}\\
Tel-Aviv, Israel 6997801 \\
sintov1@tauex.tau.ac.il}
}

\maketitle

\begin{abstract}
This paper presents a novel approach for ultra-range gesture recognition, addressing Human-Robot Interaction (HRI) challenges over extended distances. By leveraging human gestures in video data, we propose the Temporal-Spatiotemporal Fusion Network (TSFN) model that surpasses the limitations of current methods, enabling robots to understand gestures from long distances. With applications in service robots, search and rescue operations, and drone-based interactions, our approach enhances HRI in expansive environments. Experimental validation demonstrates significant advancements in gesture recognition accuracy, particularly in prolonged gesture sequences.
\end{abstract}

\begin{IEEEkeywords}
Human-Robot Collaboration, Human-Robot Interaction, Video Recognition, Ultra-Range Gesture Recognition
\end{IEEEkeywords}

\section{Introduction}
In recent years, Gesture Recognition (GR) in Human-Robot Interaction (HRI) has been advanced significantly. While research primarily focuses on hand gesture recognition \cite{ahmedhand} using conventional cameras and Kinect devices \cite{biswas2011gesture}, a critical gap remains in recognizing gestures over distances beyond seven meters \cite{nickel2007visual, wachs2011vision}. Bridging this gap is crucial for seamless collaboration between humans and robots, especially in scenarios like surveillance, drone operations, and search and rescue missions.

\begin{figure}[htbp]
\centering
\includegraphics[width=\linewidth]{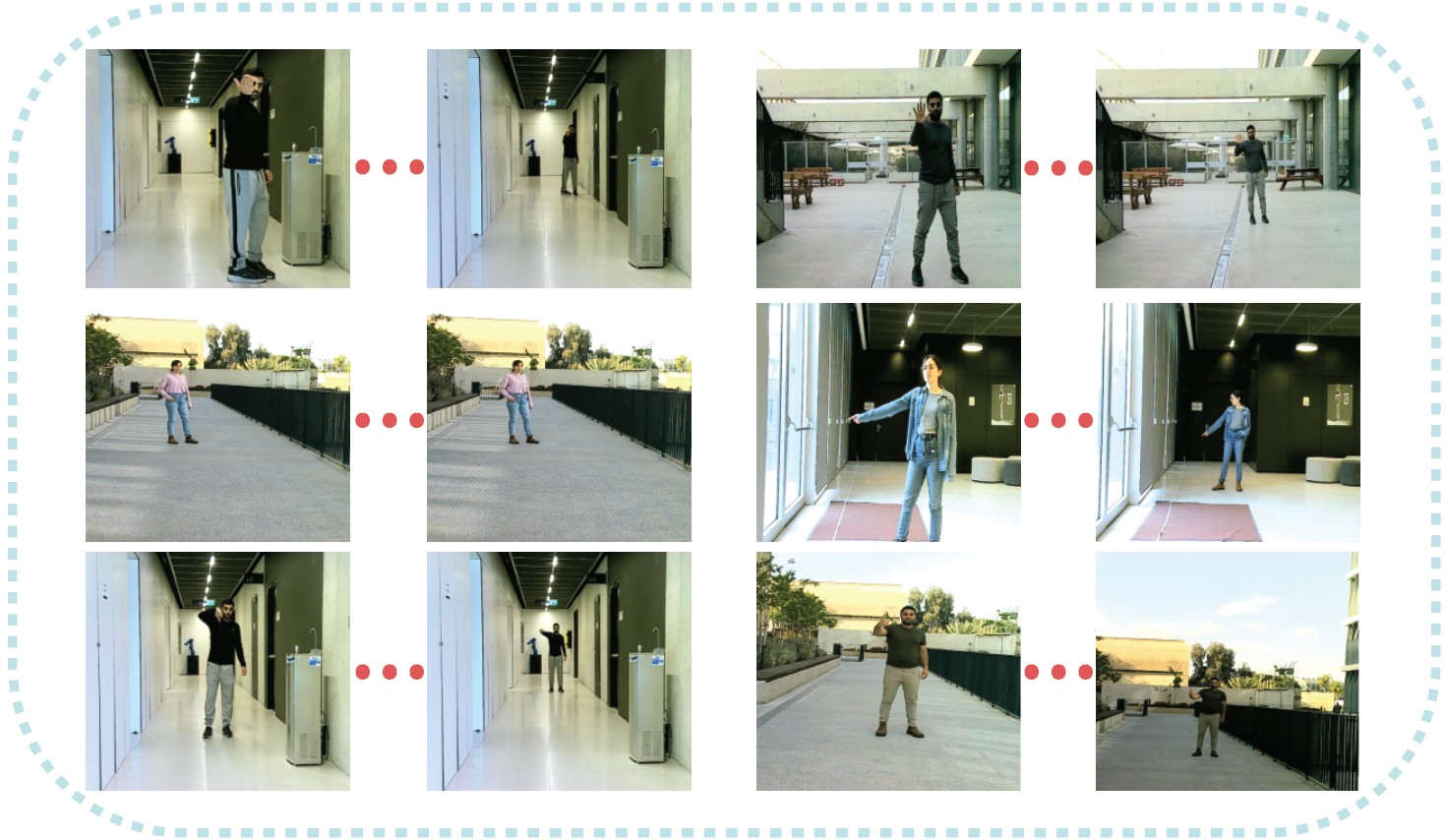}
\caption{Six human gestures are depicted with two images each, illustrating the start and end of each gesture. The gestures, arranged from the top left, include beckoning, stop, null, thumbs-up, pointing, and thumbs-down.}
\label{fig:humang}
\end{figure}

Ultra-range gesture recognition (URGR) aims to extend recognition capabilities to 25 meters \cite{bamani2024ultra}, yet considers only static images as the majority of gesture recognition approaches. However, some gestures are inherently dynamic such as waving and beckoning. Additionally, some dynamic gestures may appear as different if only a single frame is used for inference. For instance, a snapshot of a waving gesture may look like a stop gesture. In addition, URGR often faces challenges like image degradation and noise. Video data offers promise, capturing motion dynamics better \cite{materzynska2019jester, lui2012human}. Deep learning shows potential for recognizing gestures in videos \cite{yang2018temporal, bu2020human}, aiding robots in understanding gestures across distances.

In this paper, we introduce the Temporal-Spatiotemporal Fusion Network (TSFN) for URGR in HRI which is effective within a range of 28 meters and operates in real-time. The proposed TSFN model uses Temporal Convolutional Networks (TCN) \cite{lea2016temporal} for temporal dependencies and R(2+1)D 
convolutional networks \cite{tran2018closer} for spatiotemporal feature extraction, advancing ultra-range human-robot interaction.

\section{Methods}
\subsection{Problem Formulation and Data Collection}

GR across diverse environments and distances is essential for effective human-robot communication. Let $\boldsymbol{V}_i \in \mathbb{R}^{T \times H \times W \times 3}$ denote video sequences up to 28 meters, with $T$, $H$, and $W$ representing frame quantity, height and width, respectively. GR in a video faces challenges due to temporal dynamics, occlusions, and lighting variations. Given a temporal frame sequence $\boldsymbol{V} = \{v_1, v_2, ..., v_T\}$ and a set of gestures $G$, the problem entails finding the function $F: V \rightarrow G$ mapping video sequences to human gestures.

A diverse dataset $\mathcal{D} = \{(X_i, y_i, d_i)\}_{i=1}^N$ is crucial for accurate gesture recognition. It comprises input video $X_i$, gesture label $y_i$, and distance $d_i \in [4, 28]$ meters from the camera. Webcam-collected data offers varied scenarios, improving model robustness. Challenges include pose/environmental diversity and image quality degradation with distance, leading to reduced resolution and motion blur. Annotations include ground truth gesture labels and distance, aiding model training/evaluation.

\subsection{Models}

Our TSFN model combines TCN and R(2+1)D models to create a robust  GR system from video sequences. TCN captures temporal dependencies and the output of a TCN layer is given by
\vspace{-0.3cm}
\begin{equation}
Y_{\text{TCN}} = \sigma(W_{\text{TCN}} \ast V + b_{\text{TCN}}),
\label{eq:tcn}
\vspace{-0.1cm}
\end{equation}
where \( W_{\text{TCN}} \) and \( b_{\text{TCN}} \) are the weights and biases of the TCN layer, the operator \( \ast \) denotes convolution, and \( \sigma \) is the Sigmoid activation function.
Video data necessitates 3D convolutions.
A standard 3D convolution applies a filter \( W \in \mathbb{R}^{K \times d \times k \times k} \), where \( K \) is the number of output channels, \( d \) is the temporal depth, and \( k \times k \) is the spatial dimension, given by:
\[
Y_{t,h,w}^k = \sum_{c=1}^{C} \sum_{i=1}^{d} \sum_{j=1}^{k} \sum_{m=1}^{k} W_{c,i,j,m}^k \cdot V_{c, t+i, h+j, w+m}.
\]
In the TCN model, the 3D convolution is factorized into 1D temporal convolution. R(2+1)D networks extend two-dimensional convolutional operations into the temporal domain, separating spatial and temporal convolutions. The R(2+1)D convolution is expressed as
\begin{equation}
Y_{\text{R(2+1)D}} = \sigma(W_{\text{spatial}} \ast (\sigma(W_{\text{temporal}} \ast V + b_{\text{temporal}})) + b_{\text{spatial}}),
\label{eq:r2plus1d}
\end{equation}
where \( W_{\text{spatial}} \) and \( W_{\text{temporal}} \) are the spatial and temporal convolutional weights, respectively. Finally. the outputs from the TCN and R(2+1)D layers are concatenated to form combined features
\vspace{-0.2cm}
\begin{equation}
F_{\text{combined}} = [Y_{\text{TCN}}, Y_{\text{R(2+1)D}}].
\label{eq:combined_features}
\vspace{-0.1cm}
\end{equation}
These features are then passed through three fully connected layers to produce the final gesture classification:
\vspace{-0.1cm}
\begin{equation}
\text{O} = \sigma(W_{\text{fc}} \cdot F_{\text{combined}} + b_{\text{fc}}),
\vspace{-0.1cm}
\label{eq:output}
\end{equation}
where \( W_{\text{fc}} \) and \( b_{\text{fc}} \) are the weights and biases of the fully connected layer. This fusion leverages the strengths of both models to enhance GR accuracy.

To train our TSFN model to address the challenges of recognizing human gestures in a video, particularly when the subject is far from the camera, we propose a novel loss function that takes into account the degradation and other phenomena affecting the results. This novel loss function will be designed to balance classification accuracy with robustness to distance-related degradation and is given by
\[
\mathcal{L} = \mathcal{L}_{CE} + \alpha \cdot \mathcal{L}_{global} + \beta \cdot \mathcal{L}_{dist} + \gamma \cdot \mathcal{L}_{robust}
\],
where $\mathcal{L}_{CE}$ is the cross-entropy loss and $\mathcal{L}_{global}$ is the global context loss that ensures the global context of the video is captured accurately. Loss
$\mathcal{L}_{dist}$ is the distance aware loss that takes into account the degradation effect due to varying distances given by 
\vspace{-0.3cm}
\[
\mathcal{L}_{dist} = \frac{1}{N} \sum_{i=1}^{N} d_i \cdot \mathcal{L}_{CE}(v_i, y_i).
\vspace{-0.2cm}
\]
$\mathcal{L}_{dist}$ penalizes more easily for errors when the subject is farther from the camera. Robustness loss $\mathcal{L}_{robust}$ enhances the model's robustness to various degradations by adding a regularization component that minimizes the variance in predictions due to changes in distance:
\[
\vspace{-0.3cm}
\mathcal{L}_{robust} = \frac{1}{N} \sum_{i=1}^{N} \sum_{j=1}^{M} (\mathcal{L}_{CE}(v_{i}^{(j)}, y_{i}) - \bar{\mathcal{L}}_{CE}(y_{i}))^2.
\vspace{-0.1cm}
\]
Component $\bar{\mathcal{L}}_{CE}$ is the mean cross-entropy loss for the true label, $\alpha, \beta, \gamma$ are hyper-parameters that control the relative importance of each loss component.

\section{Experimental Results}

\begin{figure}[htbp]
\centering
\includegraphics[width=\linewidth]{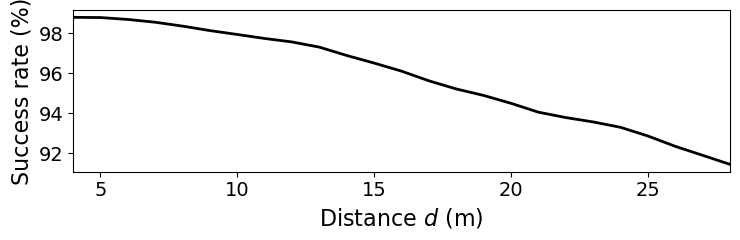}
\vspace{-0.7cm}
\caption{Model performance vs. distance. The plot shows the accuracy of the TSFN model in recognizing gestures at various distances.}
\label{fig:performance_vs_distance}
\end{figure}

To validate the effectiveness of the proposed TSFN model, we conducted experiments comparing its performance with state-of-the-art video recognition models. Figure \ref{fig:humang} illustrates the six gesture categories we focused on: pointing, thumbs-up, thumbs-down, beckoning, stop, and null gestures. The models were tested on a dataset of human gestures observed from 4 to 28 meters. We collected 15 samples of 4-second videos per meter, totaling 2,250 videos for each gesture, with 150 videos used as the test set. Key performance metrics including accuracy, loss, and Mean Average Precision (mAP) are shown in Table \ref{tab1}. Figure \ref{fig:performance_vs_distance} shows the accuracy of the TSFN in GR with respect to different distances. The results clearly show the superiority of TSFN over existing methods.

\begin{table}[htbp]
\caption{Comparison of Video Recognition Models}
\begin{center}
\begin{tabular}{lcccc}
\toprule
\textbf{Model} & & \textbf{Accuracy (\%)} & \textbf{Loss} & \textbf{mAP} \\
\midrule
ViViT & \cite{arnab2021vivit} & 91.2 & 0.29 & 0.87 \\
TCN & \cite{lea2016temporal} & 89.3 & 0.37 & 0.79 \\
R(2+1)D & \cite{tran2018closer} & 88.1 & 0.39 & 0.77 \\
Vanilla CNN & & 78.4 & 0.51 & 0.70  \\
CNN+LSTM & \cite{chen2023cnn} & 83.7 & 0.42 & 0.74  \\
\rowcolor{lightgray}
TSFN & & 96.1 & 0.12 & 0.92  \\
\bottomrule
\end{tabular}
\label{tab1}
\end{center}
\end{table}

\section{Conclusion}
TSFN redefines ultra-range gesture recognition, which is crucial for seamless Human-Robot Collaboration. By fusing TCNs and R(2+1)D networks, TSFN excels in understanding gestures over vast distances. Future work includes refining TSFN for real-time applications and exploring its potential in dynamic environments including underwater.

\end{document}